\documentclass{article}
\usepackage{ijcai25}

\usepackage{times}
\usepackage{soul}
\usepackage{url}
\usepackage[hidelinks]{hyperref}
\usepackage[utf8]{inputenc}
\usepackage[small]{caption}
\usepackage{graphicx}
\usepackage{amsmath}
\usepackage{amsthm}
\usepackage{booktabs}
\usepackage{algorithm}
\usepackage{algorithmic}
\usepackage[switch]{lineno}
\usepackage[dvipsnames]{xcolor}

\title{Efficient Mixture of Geographical Species for On Device Wildlife Monitoring}

\author{
Emmanuel Azuh Mensah$^1$
\and
Joban Mand $^1$\and
Yueheng Ou$^{1}$\and
Min Jang$^1$ \And Kurtis Heimerl$^{1}$\\
\affiliations
$^1$University of Washington\\
\emails
\{emazuh, jmand1, yueheng5, minjang, kheimerl\}@cs.washington.edu
}

\begin{document}

\maketitle

\begin{abstract}
Efficient on-device models have become attractive for near-sensor insight generation, of particular interest to the ecological conservation community. For this reason, deep learning researchers are proposing more approaches to develop lower compute models. However, since vision transformers are very new to the edge use case, there are still unexplored approaches, most notably conditional execution of subnetworks based on input data. In this work, we explore the training of a single species detector which uses conditional computation to bias structured sub networks in a geographically-aware manner. We propose a method for pruning the expert model per location and demonstrate conditional computation performance on two geographically distributed datasets: iNaturalist and iWildcam.
\end{abstract}

\section{Introduction}


The widespread success of transformer models has permeated many applications, now including resource constrained use cases such as improving consumer experience and for near sensor intelligence. This interest in reducing the cost of downstream deployment and finetuning has spurred many energy efficient algorithms for deep (transformer) models \cite{tang2024survey}. These approaches have included quantization for efficient model weight storage, pruning of low importance parameters and low rank decomposition of weight matrices  \cite{gholami2022survey}. 
However, input dependent subnetwork computation still remains under-explored for edge vision transformers. Although there have been recent efforts to find subnetworks that specialize on a subset of in-domain classes as a means to reduce the deployed model size, the resulting model doesn't dynamically run a subset of the network layers depending on input images \cite{kuznedelev2023vision}. We seek to investigate a simple but important question: ``can we train edge vision transformers with the intention of selecting class-based subnetworks, possibly without finetuning''? More specifically, our interest in the ecological monitoring use case leads us to investigate whether we can use spatial biases in species distribution to train geography aware conditional subnetworks in edge vision transformers as an approach to structured sparsity during inference. 

\begin{figure*}[ht]
  \centering
   \includegraphics[scale=0.41]{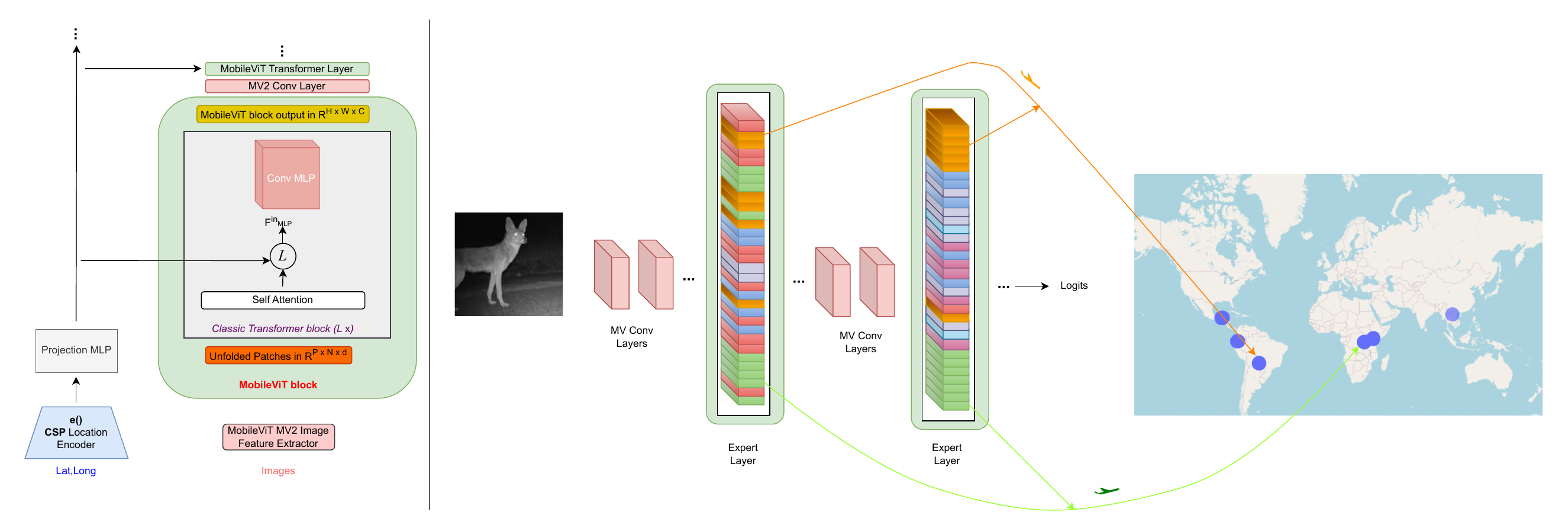}
   \caption{The proposed model setup first finetunes MobileViTV2 \protect\cite{mehta2022_mvitv2}, with geographical location embeddings from \protect\cite{wu2024torchspatial} corresponding to each image, which is included in a multimodal supervised contrastive formulation ($\mathcal{L}$) before each transformer layer MLP. We freeze the location encoder but finetune the location projection MLP. After fine tuning, we follow the same approach as \protect\cite{mensah2024visionmixtureexpertswildlife} to derive the experts model. Experts not activated for a deployment location can then be pruned away.}
   \label{fig:system_diagram}
\end{figure*}

\subsection{Motivation for Sub-Networks}
In many application areas such as ecological monitoring, end users who typically have low or no compute budgets can benefit from reduced burden of finetuning. Even with tools like Google Colaboraory \cite{bisong2019google} which provides users an option to finetune custom models, there is still a need to reduce the expertise required for finetuning and to reduce the combined energy utilization over many users. Especially in scenarios where machine learning is used to support sustainability, the need for sustainable machine learning is even greater. In addition, for smaller companies running edge ML for internet of things, they will have to finetune a new set of downstream sub-models whenever the upstream model is updated, for instance due to newly collected data \cite{ding2024lora}. Reducing the burden for finetuning by training with data group based subnetworks in mind would be useful for easier edge model updates.

\subsection{Geography Based Data Groups}
Location information has been used as a way to constrain learning to improve performance when spatial biases exist in datasets such as species and land cover datasets \cite{wu2024torchspatial}. 
We therefore hypothesize that mixture-of-expert subnetworks could be trained with such prior supervision. This would then lead to specialized subnetworks useful for categorizing species for various locations that have different distribution of observed species. We test our hypothesis with two species datasets with geographical location metadata available - iNaturalist (community science) \cite{van2018inaturalist} and iWildcam (camera trap) \cite{koh2021wilds}. Our contributions are as follows:


\begin{enumerate}
    \item A training proposal for incorporating GPS information associated with species image datasets in the process of creating geography aware Mixture of Experts models.
    \item An efficient method for quickly computing the subset of experts in the MoE model that are most important for classifying species from a geographic location and analyze when it breaks down.
    \item Evaluation of our method on two geographically distributed species datasets demonstrating the possibility to train a single model on large scale geo-data and have users in specific locations download a subset of the model.
\end{enumerate}

\section{Related Work}

As the internet and sensor technology continue to improve global data access, adding location information tends to improve performance for machine learning models that operate on global datasets \cite{wu2024torchspatial}. With support steadily improving from the data collection infrastructure, to algorithm design and deployment processes, the ecosystem of tools and algorithms are still under development for conservation use cases and can benefit from further improvement.


\subsection{(Camera Trap) Species Identification} 
Camera traps have become a popular tool for monitoring rare and endangered species \cite{chalmers2019conservation}, studying ecosystems 
and keeping track of environmental patterns such as height of snow cover \cite{strickfaden2023virtual} in remote locations. The rise of camera trapping has led to the development of machine learning algorithms for automated analysis of large quantities of sequential image data. These algorithms are used for species identification \cite{norouzzadeh2018automatically},
animal re-identification \cite{schneider2020similarity}, and removal of empty images \cite{beery2018recognition}. 
In spite of the increase in the number of tools and infrastructure to support machine learning for ecological use cases, there is still a gap in expertise for ecologists, leading ecology researchers with machine learning experience to develop online training communities and share model training guidelines \cite{blair2024gentle}. This is because the tools tend to be generalists and mostly developed for post-processing data. For real-time monitoring scenarios where near-sensor identification \cite{tuia2022perspectives} is needed  for a subset of species in a generalist model, one has to set up a full training pipeline and rely on trial and error.
Recent work such as \cite{kuznedelev2023vision} aims to close this gap by developing a general-purpose algorithm that takes a generalist model and a small training dataset that is within the domain (or moderately out of domain) of the generalist model’s training data. They then use linear solvers to find a data conditional subnetwork that is typically smaller than the generalist model.


\subsection{Geographical Prior Training} 
Geo aware classification has been successfully applied to data collected alongside GPS coordinates such as geotagged images from Flicker Photos \cite{spyrou2016analyzing} and iNaturalist Application \cite{mac2019presence}, Land Use datasets \cite{yang2010bag} and remote sensing satellite images \cite{sumbul2019bigearthnet}. Location information can be utilized in an unsupervised or self supervised setup such as using location as an anchor to select a co-occurring image and an image from a distant location to form a triplet loss \cite{jean2019tile2vec}, and using location as an anchor to select images occurring in the same place at different times as a positive pair \cite{ayush2021geography}. For the supervised setting, location can be grouped into bins for a pretext training formulation where the supervision signal is to predict which bin an image falls in, typically with a cross-entropy loss \cite{zhai2019learning}. More recently, spatially explicit artificial intelligence has been proposed to utilize more inductive biases in location information such as spatial continuity,
temporal periodicity 
and the earth’s spherical geometry \cite{mai2023csp}.
In this work, we are interested in how the learned intermediate layer embeddings from spatially explicit AI can be used to generate spatially sensitive conditional subnetworks.

\subsection{Energy Efficient Deployment for Vision Transformers} 
The most popular approaches to improving vision transformer model efficiency for deployment include quantization \cite{gholami2022survey}
, knowledge distillation ~\cite{touvron2021training},
efficient modules within deep models \cite{hatamizadeh2023fastervit}
 and pruning \cite{luo2017thinet}.
The most general of these approaches are quantization and pruning based on unstructured sparsity since these approaches make minimal assumptions about model weight and activation distributions. Quantization has gained a lot of community traction and many edge devices support low bit representation \cite{oruacsan2022benchmarking}
. Knowledge distillation and deep module design tend to require availability of a more complex model to distill, larger datasets or expertise of specific model architectures. From the hardware architecture perspective, structured sparsity tends to be slightly more favored since it leads to lower computational complexity \cite{hubara2021accelerated}. We consider data dependent structure of weights and activations for the under explored  benefits of structured sparsity for edge vision transformers.

\section{Design}
In this section, we describe an approach to include geographical bias during training in order to segment subnetworks sensitive to species and geographical contexts. 

The standard mobile vision transformer model design \cite{mehta2022_mvitv2} alternates between MobilenetV2 inverted residual convolution blocks \cite{sandler2018mobilenetv2} and transformer blocks \cite{shazeer17_moe}. As in the mixture-of-experts (MoE) literature, we focus on the Multi-Layer Perceptron (MLP) of the transformer block for the conditional subnetwork \cite{shazeer17_moe}. Since the embedding of the input to the MLP block determines the segmentation of experts in an MoE block, we include the geographical information in generating location-aware experts in three steps: i) Finetune a MobileViTV2 model on a species dataset detailed in section \ref{sec:dataset}. ii) Include embeddings from the location encoder detailed in \ref{sec:loc_enc} to the pre-MLP activations in a contrastive setting. iii) Generate the expert gate and fine-tune the expert model according to \cite{mensah2024visionmixtureexpertswildlife}. After these steps, we can then quantify the importance of each expert MLP on the classification of a non-zero set of species of interest. MLPs can the be pruned based on importance when deploying to a location with the species of interest.



\subsection{Globally Distributed Species Dataset}
\label{sec:dataset}
We consider two major datasets to test our proposed model in the context of global species monitoring: i) iNaturalist with dense instances of citizen science species data and ii) iWildcam with sparse instances of camera trap data.

\begin{figure}[!tb]
\centering
    \includegraphics[width=0.7\linewidth]{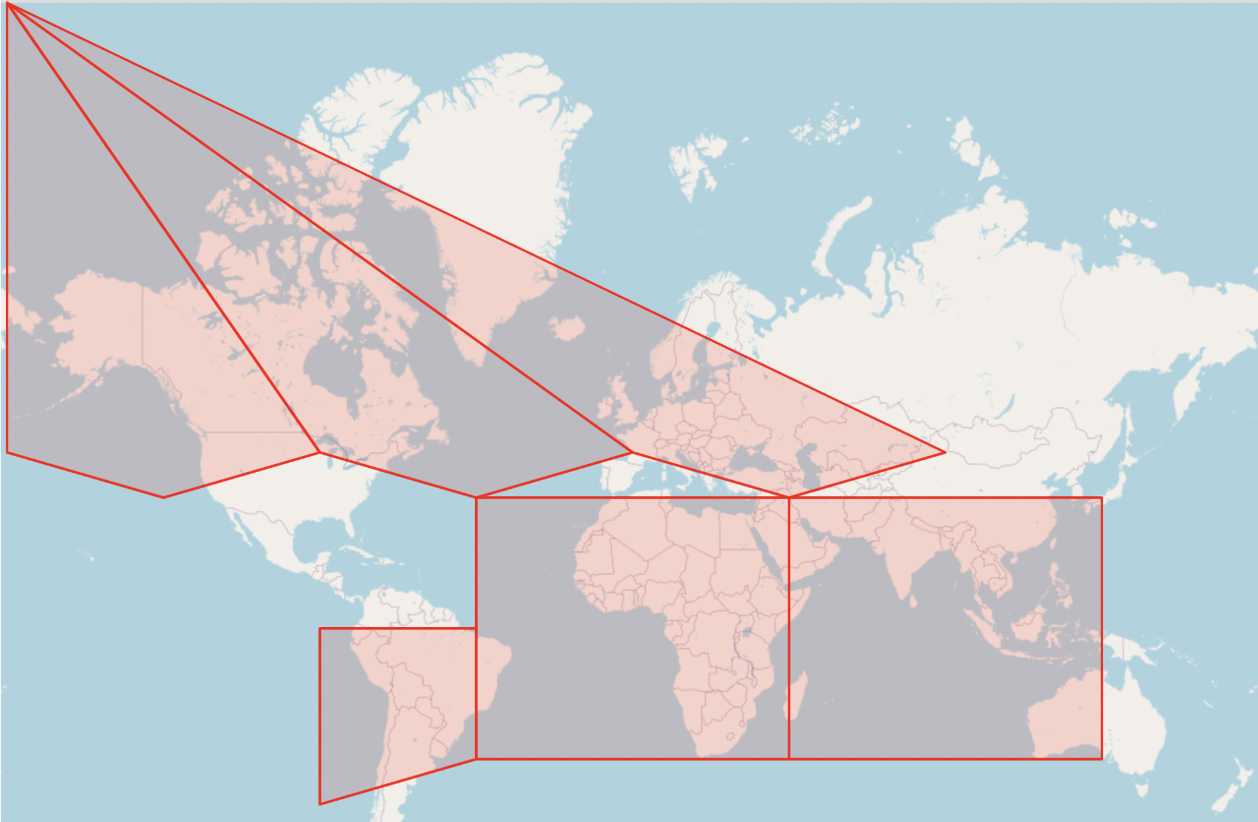}

   \caption{S2 regions for the 6 identifiers `2/1', `2/2', `2/0', `0/', `1/', and `4/2', used in evaluating the effect of location grouped species on subnetwork importance for iNaturalist-Geo-10K.}
   \label{fig:s2_10k}
\end{figure}

\textbf{iNaturalist}
The inaturalist dataset is the largest citizen science dataset in the world. Participants around the globe upload images of species they encounter during outdoor activities and experts curate standardized datasets from the submissions. For this reason, there is typically a geographical bias towards richer countries and areas with foot traffic. Regions in Africa, South America and Asia with rich biodiversity are underrepresented in this dataset. However, the data provides very fine grained location information over a broad area globally. We use the iNaturalist 2017 dataset \cite{van2018inaturalist} in this paper. In addition to the full iNaturalist dataset, we also use \textbf{iNaturalist-Geo-$N_{max}$} proposed by \cite{hsu2020federated}. This subset of iNaturalist is split with class balance in mind. Using the S2 global addressing system based on Hilbert Curves, one can generate unique identifiers for grids in a hierarchy of resolutions \cite{weyand2016planet}. \cite{hsu2020federated} varies the hierarchy level so that the maximum number of data instances for any species is between $N_{max}$ and $0.1N_{max}$ within a grid. We use $N_{max}=10,000$ with sample S2 cells splits shown in Figure \ref{fig:s2_10k}. 

\textbf{iWildCam}
This dataset is closest to our use case of interest in that it features images from camera traps installed in 552 camera locations from 12 countries in different parts of the globe \cite{beery2021iwildcam}. Although this dataset is more globally represented than most other camera trap datasets, the individual locations are tightly clustered to small conservation zones selected by the Wildlife Conservation Society leading to a coarse grained representation of species in global regions. All species in iWildcam are also present in iNaturalist. We use the \textbf{iWildcam2020-WILDS} dataset presented in WILDS benchmark \cite{koh2021wilds}. After filtering for reliability (e.g. removing empty frames) from the dataset, only $182$ severely imbalanced classes in 5 continents remain.

\begin{figure}[h]
\centering
    \includegraphics[width=0.7\linewidth]{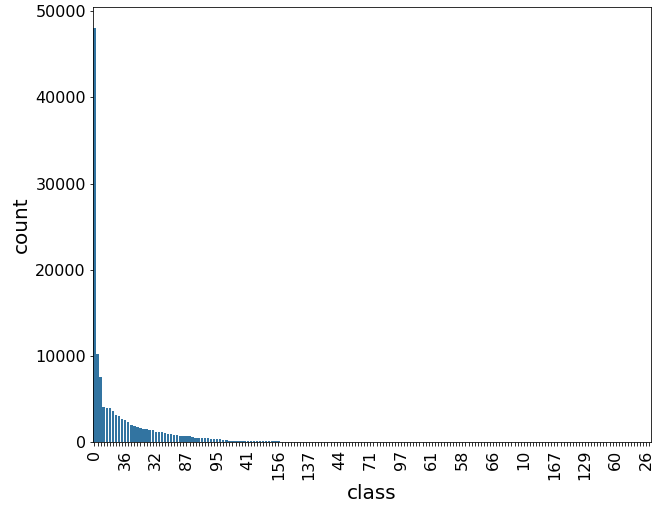}

   \caption{\small Frequency plot of iWildcam2020-WILS dataset, demonstrating the difficulty of obtaining class balanced camera trap datasets to develop models that generalize well at deployment time.}
   \label{fig:iwild_imabalance}
\end{figure}

Table \ref{tab:data_split_sizes} shows the split of the two dataset across 6 locations chosen to cover distinct geographical regions as shown in Figure \ref{fig:s2_10k} for iNaturalist and Figure \ref{fig:system_diagram} (right) for iWildcam. The test sets are filtered to include only species found in the training data for the locations of interest.

\begin{table}[!htbp]
    \centering
    \scriptsize
    \begin{tabular}{cccccccc} \\ 
    \hline
    Dataset & Split & L1 & L2 & L3 & L4 & L5 & L6 \\
    \hline
    iNat-10K & Train & 7012 & 5293 & 3640 & 1481 & 556 & 217 \\
    & Val & 1754 & 1324 & 910 & 371 & 139 & 55 \\
    & Test & 16573 & 18675 & 6560 & 10667 & 2468 & 2088 \\
    \hline
    WILDS & Train & 27199 & 5405 & 6618 & 598 & 89 & 6196 \\
    & Val & 2401 & 373 & 723 & 27 & 22 & 513 \\
    & Test & 7905 & 5110 & 4152 & 3662 & 2132 & 7380 \\
    \hline
    \end{tabular}
    \caption{Data points for various locations L1 to L6 of the datasets used for evaluating the proposed geo aware model.}
    \label{tab:data_split_sizes}
\end{table}

\subsubsection{Global Spread of Species}
\begin{figure}[!htb]
  \centering
   \includegraphics[width=\linewidth]{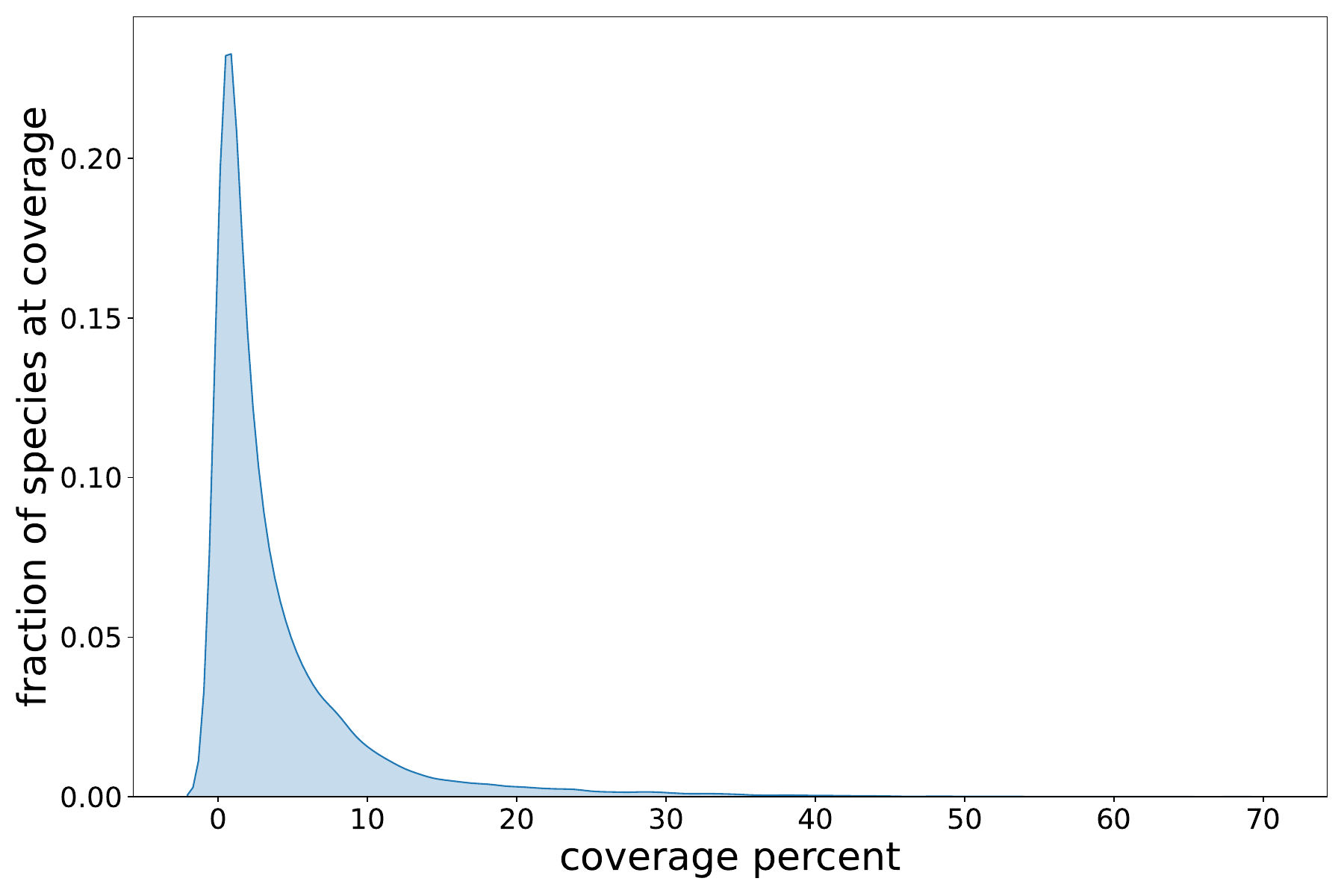}

   \caption{Density plot of global coverage percentage of various species. iNaturalist species distribution prediction data suggests that many species have local distribution, a characteristic useful in designing a geographical based mixture of experts model.}
   \label{fig:species_dist}
\end{figure}

Our model proposal relies on the assumption that there are enough species that are confined to small regions of the world. If this assumption doesn't hold, we would expect to have all conditional expert MLPs to be needed for every location. To investigate the species spread for iNaturalist 2017 dataset \cite{van2018inaturalist}, we obtain a presence-only species distribution coverage on land using a pretrained model provided by ~\cite{cole2023spatial}. As seen in Figure \ref{fig:species_dist}, the coverage percent of iNaturalist species follows a power law distribution which means very few species can be found in all continents and most species are confined to small spatial regions. The distribution plot can be updated over time as more species are reported to iNaturalist website, in order derive more confident distribution information.



\subsection{Encoding Location}
\label{sec:loc_enc}
We use the Space2Vec-grid location encoder $e_{\theta}(x_i):S^2 \rightarrow S^d$ proposed in \cite{mai2020multi} for location encoding. Given a longitude ($\lambda_i \in [-\pi,\pi]$), latitude ($\phi_i \in [-\pi/2, \pi/2]$) pair $x_i = (\lambda_i, \phi_i)$ in a spherical surface $S^2$, $e_{\theta}$ returns a $d=256$ embedding vector.
We train the geography aware species model with a contrastive pre-trained Space2Vec-grid model for iNaturalist 2017 provided by the TorchSpatial framework \cite{wu2024torchspatial}. 
Given an Image, Location pair ($I, x$), the pre-MLP image feature representation in a MobileViT transformer layer is given by 
$$
F^{in}_{MLP} = Norm(MHSA(E_I) + E_I)
$$ 

where $E_I$ is a embedding of the image at the input of the transformer layer, $Norm$ is a layer normalization operation and $MHSA$ is a multihead self attention layer. The location embedding at a transformer layer is represented as

$$
F_{loc} = Dropout(\sigma(MLP^{b}_{proj}(Norm(e(x)))))
$$

where $e$ is the location encoder from TorchSpatial \cite{wu2024torchspatial}. $MLP^{b}_{proj}$ is a single layer perceptron that projects the $R^{256}$ location embedding to the transformer block $b$ with dimension $64$, $96$ and $128$ respectively for the three transformer blocks in MobileViTV2-0.5. Each transformer layer within the same block uses the same $MLP^{b}_{proj}$. We use ReLu for the activation ($\sigma$). $F^{in}_{MLP}$ and $F_{loc}$ thus represent the multiview instance pair utilized in the supervised contrastive loss formulation $\mathcal{L}^{sup}_{out}$
 from \cite{khosla2020supervised} to force embeddings of images and locations within the same class to be close to each other. We compute this contrastive loss on each of the 9 transformer pre-MLP activations (from transformer layers 0-8), weighted by a constant $0.01$ as an auxiliary loss to be added to the classification loss. To reduce overfitting from the location encoding, we use a dropout rate of $0.3$. 

\begin{figure}[!b]
  \centering
   \includegraphics[width=.7\linewidth]{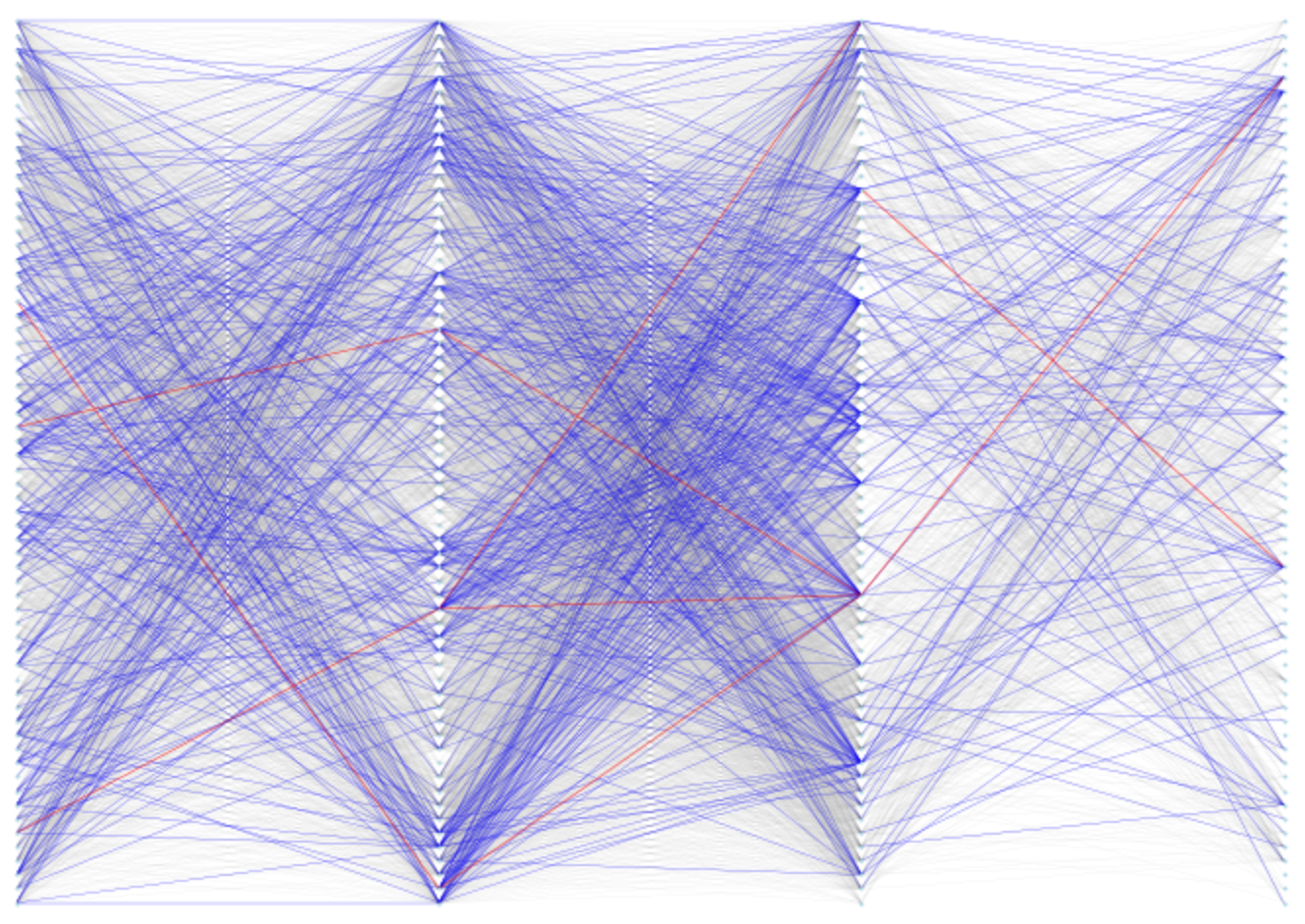}

   \caption{Sample cross layer routing for S2 cell `2/1' at layers $1,3,5,7$ of a $64$ expert MoE model. Grey lines represent routes in the bottom 90th percentile utilization, blue lines are between 90th and 99.9th percentile and red lines are above 99.9th percentile. 
   }
   \label{fig:s2_cross}
\end{figure}


\begin{figure*}[tb!]
\begin{center}
    \begin{tabular}{lllll}
        \includegraphics[width=.16\linewidth]{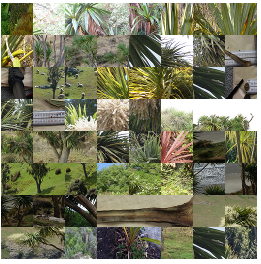} & \includegraphics[width=.16\linewidth]{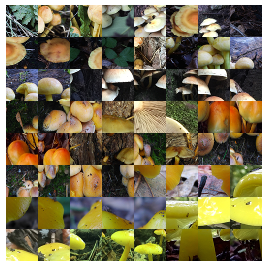} & \includegraphics[width=.16\linewidth]{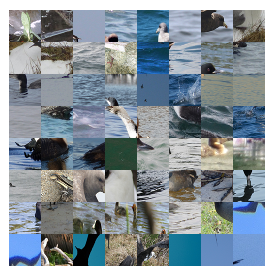} & \includegraphics[width=.16\linewidth]{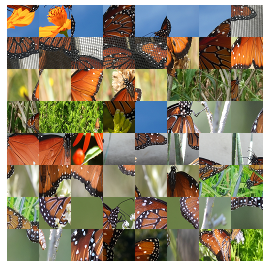} & \includegraphics[width=.16\linewidth]{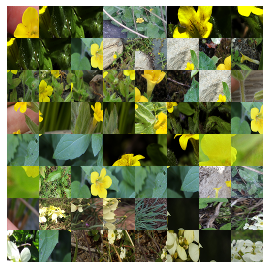} \\
        \includegraphics[width=.16\linewidth]{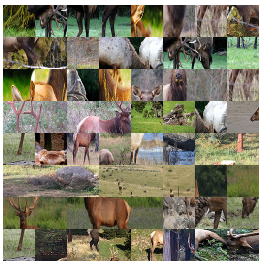} & \includegraphics[width=.16\linewidth]{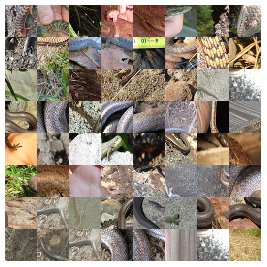} & \includegraphics[width=.16\linewidth]{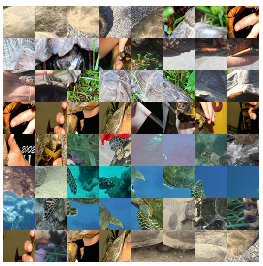} & \includegraphics[width=.16\linewidth]{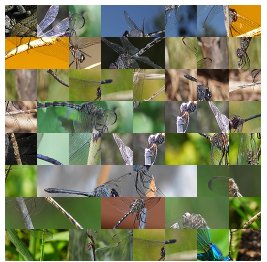} & \includegraphics[width=.16\linewidth]{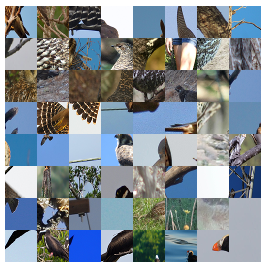} \\
        \\
        
        \includegraphics[width=.16\linewidth]{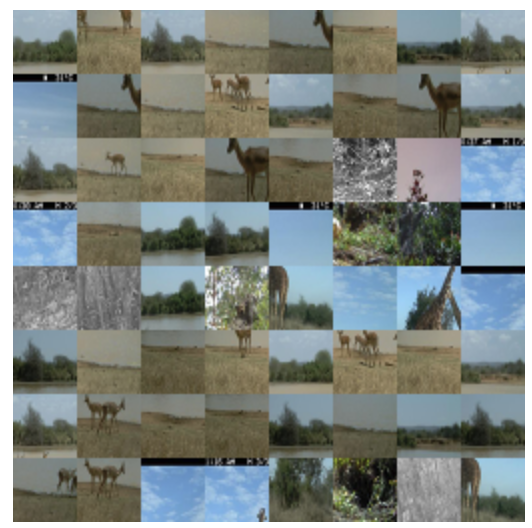} & \includegraphics[width=.16\linewidth]{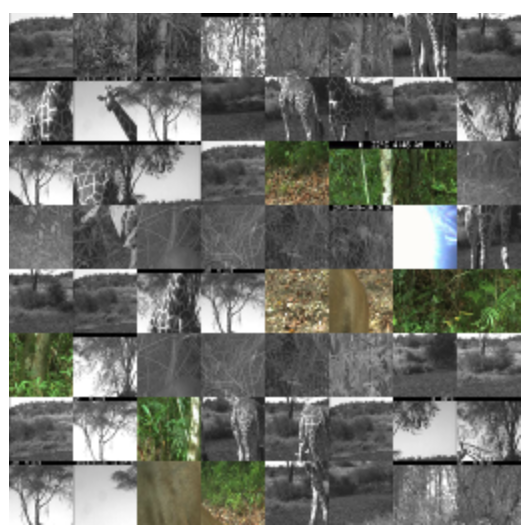} & \includegraphics[width=.16\linewidth]{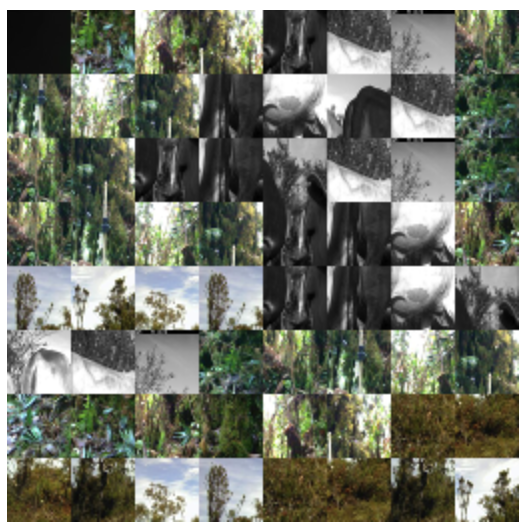} & \includegraphics[width=.16\linewidth]{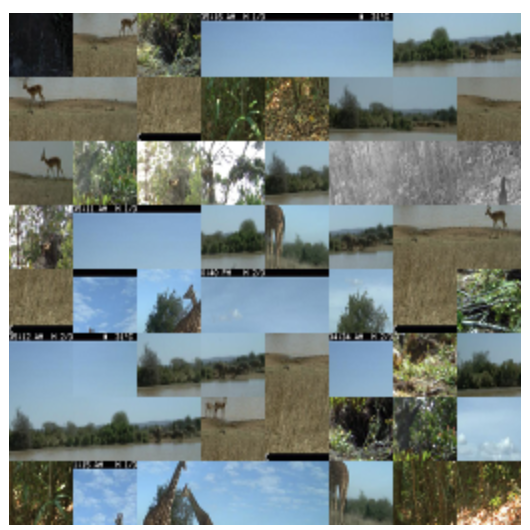} & \includegraphics[width=.16\linewidth]{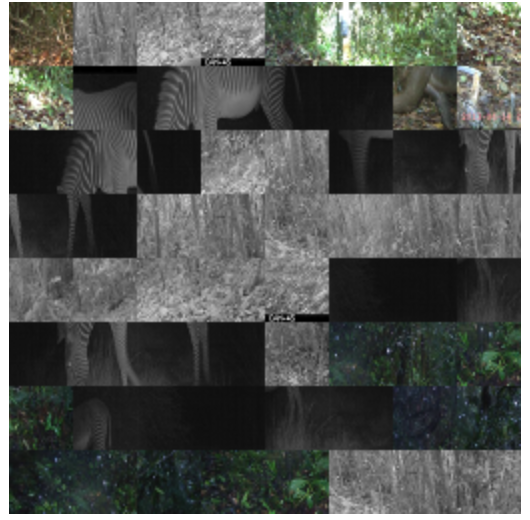} \\
        \includegraphics[width=.16\linewidth]{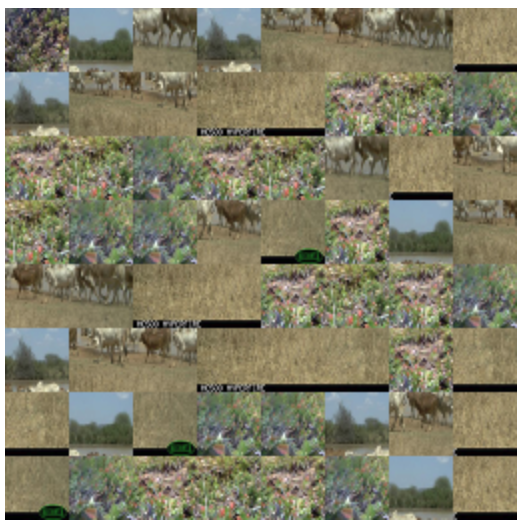} & \includegraphics[width=.16\linewidth]{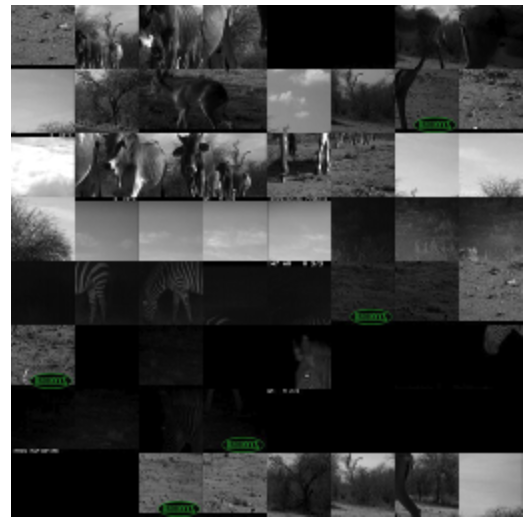} & \includegraphics[width=.16\linewidth]{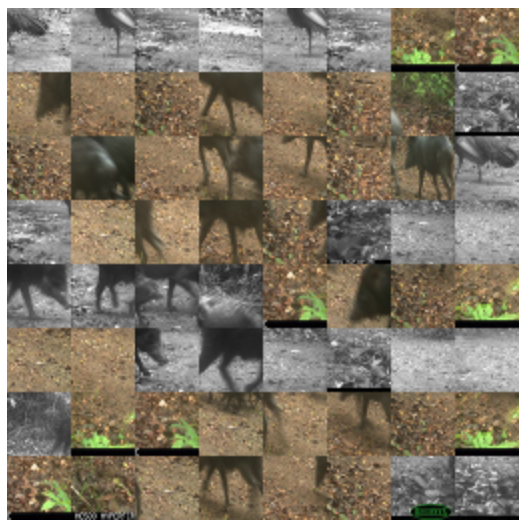} & \includegraphics[width=.16\linewidth]{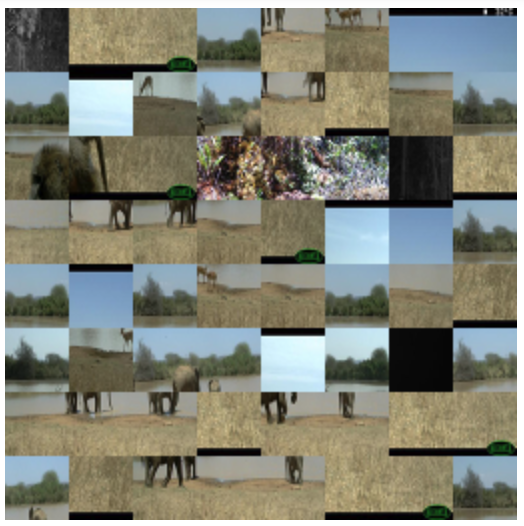} & \includegraphics[width=.16\linewidth]{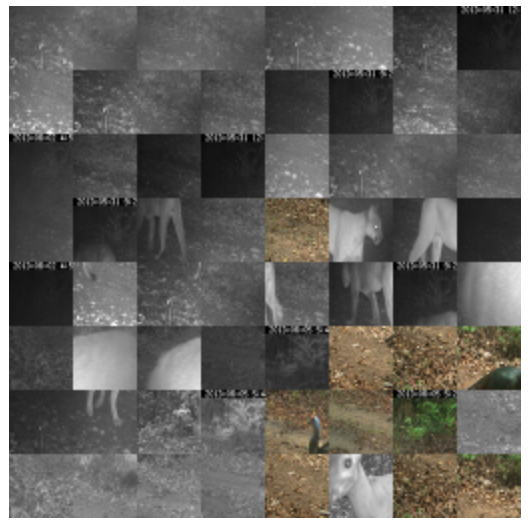} \\
        
    \end{tabular}

    \caption{ Expert groupings in the last MobileVitV2 transformer layer for iNaturalist 2017 (top 2 rows) and WILDS-iWildcam (lower 2 rows) demonstrating patches from species typically encountered in ecological monitoring settings. We report the closest patches to the centroid for every 8th expert using an expert gate initialized with 64 cluster centroids obtained from kmeans grouping of training data embeddings.}
    \label{fig:inat17_sample_semantic_split}
\end{center}
\end{figure*}




\subsection{Generating Experts and Quantifying Importance}
\label{sec:threshold_dropping}

Once a location aware model is trained according to Section \ref{sec:loc_enc}, we follow the approach in \cite{mensah2024visionmixtureexpertswildlife} to generate activation clusters at each pre-MLP layer in order to initialize the expert router (gate). In this case, the gate is initialized with location aware embeddings. 

\subsubsection{Dealing with Class Imbalance during Expert Initialization.} 
For a highly class imbalanced dataset like iWildcam, where some classes have a single datapoint as can be seen from Figure \ref{fig:iwild_imabalance}, underrepresented classes have no influence on the clustering used to initialize the router. Specifically for iWildcam, this issue worsens due to the abundance of background patches. We take two steps to improve iWildcam router initialization (i) resample data for classes with few images so all classes have a minimum of 10 images (ii) increasing the number of refinement steps proposed in \cite{mensah2024visionmixtureexpertswildlife} to $100$ as compared to $5$ for iNaturalist to improve representation of foreground patches. 

\subsubsection{Expert Importance.}
In order to compute the importance of all experts for a given geographical location (an S2 cell for federated iNaturalist or camera cluster for iwildcam), we first filter the test set to only include species found in that location. We use this filtered set of species in the validation dataset to generate a patch routing map through the model as shown in Figure \ref{fig:s2_cross}. For each (outgoing) layer, we count the number of patches using each edge and normalize by the sum of patches that went through that layer. Figure \ref{fig:s2_cross} shows a that few experts are used very often. We use experts along less trodden routes as a proxy for low importance experts. This way, we can compute the accuracy drop at once for all experts in the model that are used above a certain threshold by the filtered validation set. For a node $v$ in the patch routing graph, if all edges $E(V, \{v\})$ that start or end at $v$ have edge weights less than a threshold (e.g. 90th percentile of layer normalized edge weights), then we consider $v$ less important and marked for removal. 
We simulate expert removal by $MLP_{Exp}(F^{in}_{MLP}) \leftarrow -F^{in}_{MLP}$. Since the residual of $F^{in}_{MLP}$ is added to the output of the expert layer, the effective representation of the patch becomes a vector of zeros. This percentile threshold approach is a heuristic that avoids computing the expert importance over the entire filtered dataset for each expert and is therefore an underestimate of the filtered expert set importance estimated after computing per-expert importance (shown in Figure \ref{fig:exp_drop_method}). A sample computation of location level importance is shown in Figure \ref{fig:geo_exp_drop}. As the pruning percentile threshold increases, more experts are dropped which in turn reduces accuracy.

\begin{table*}[!htbp]
    \centering
    \scriptsize
    \begin{tabular}{lrrrrr} \\ 
    \hline
    
      Model & Params (M) & iNaturalist2017 & iNaturalist-Geo-10K & iWildcam ID & iWildcam OOD  \\
     \hline
     FLYP \cite{goyal2023finetune} & 428 & - & - & 76.2 & 76.2 \\
     PTFL \cite{chen2022importance} & 13.78 & - & 46.6 & - & - \\
     GNFG ~\cite{chu2019geo} & 9 & 72.23 & - & - & - \\
     \hline
     Mobilenet & 8.74 & 59.61 & 61.19 & 60.53 & 51.47 \\
     
     
     
     MobileViTV2-0.5 & 2.94 & 60.77 & 65.58 & 69.13 &  67.66  \\
     
     
     MobileViTV2-loc-all & & 64.55 & 64.45 & 70.94 & 63.22 \\
     MobileViTV2-loc-last-two &  & 64.51 & 65.78 & 
     70.32 &  63.02 \\
     
     
    \hline
    
    \end{tabular}
    \caption{Baseline results of training iNaturalist and iWildcam with location supervsion. \textbf{loc-\{all,last-two\}} refers to contrastive learning with location information in all MobileViTV2-0.5 transformer blocks or in the last two blocks respectively. iWildcam test set are split between In Distribution (ID) and Out of Distribution (OOD).}
    \label{tab:geo_loc_models}
\end{table*}

\subsection{Training Details}
We finetune imagenet initialized models on inaturalist and iWildcam ahead of expert conversion, using input image resolution of $224\times224$. We train all models using AdamW optimizer with default parameters, weight decay of $1e-8$, and batch size of 256. All training and finetuning experiments are ran for $60$ epochs. Input data augmentation consisted of random cut, Pytorch RandAugment as suggested by \cite{cunha2023bag}, and random flip with probability $0.5$. We also normalize images to the range of $0-1$ for MobileViT based models and Imagenet per channel average for mobilenet. All experiments use mixup of $0.1$, dropout in classifier head of $0.1$ and label smoothing $0.05$. We did not use class balance sampling during training. MobileViTV2 models are trained with a learning rate of $1e-4$. MoE models use a learning rate of $1e-3$ for expert MLPs, $1e-6$ for the classifier head and $1e-4$ for the rest of the model, using warmup and final learning rate as $0.8$ of peak learning rate with a cosine scheduler and $40$ warmup epochs.

The classifier head for the MobileViTV2 and derivative expert models for iNaturalist dataset first projects the $R^{256}$ pre-classifier encoder feature vector to $R^{1024}$, followed by a bottleneck layer of $R^{256}$ and finally $R^{5089}$ class units. This set up provides a more complex classifier that is comparable to the MobilenetV2 classification head of dimension $R^{1280\times5089}$ but with significantly fewer parameters ($1.8M$ compared to $6.5M$). We also reduce the weight of the expert gate matrix through low rank approximation with rank $8$.
\section{Evaluation}


        
    
In this section, we analyze the effects of adding location to the model inference dynamics, starting from input data grouping to effects on model resource requirements and classification accuracy on the datasets.






\subsection{Expert Patch Assignment}

\begin{figure}[!b]
\centering
    \includegraphics[width=0.9\linewidth]{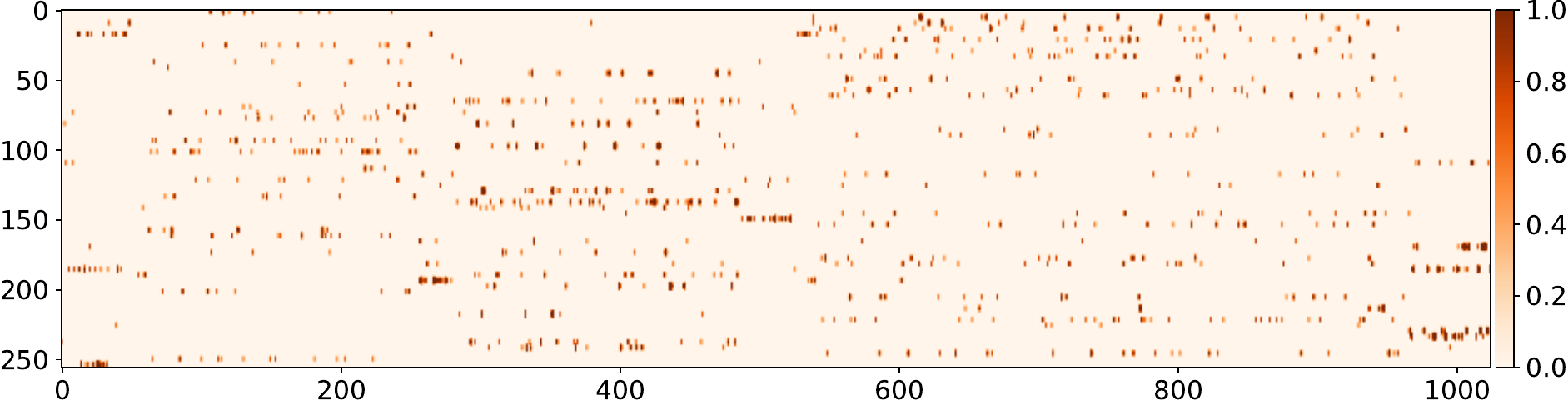} 

   \caption{\small Affinity plot for inaturalist classes (x-axis) to 64 experts (y-axis). The dimension of the y-axis is scaled up by a factor of 4 and the x-axis skips every 5 classes ($~1000$ out of $5089$) for visibility. 
   }
   \label{fig:inat_every_5_affinity}
\end{figure}

Figure \ref{fig:inat17_sample_semantic_split} shows a sample assignment of species patches to various experts at initialization. Figure \ref{fig:inat_every_5_affinity} also includes a sample affinity of patches from 1000 classes in iNaturalist 2017 to 64 experts (skipping every 5 classes out of 5098 after sorting class labels). The affinities are computed by assigning each patch from selected images to an expert based on the router initialization with Softmax temperature $0.001$ and then normalizing by the number of patches in a class per expert.

\begin{figure}[!t]
  \centering
   \includegraphics[width=\linewidth]{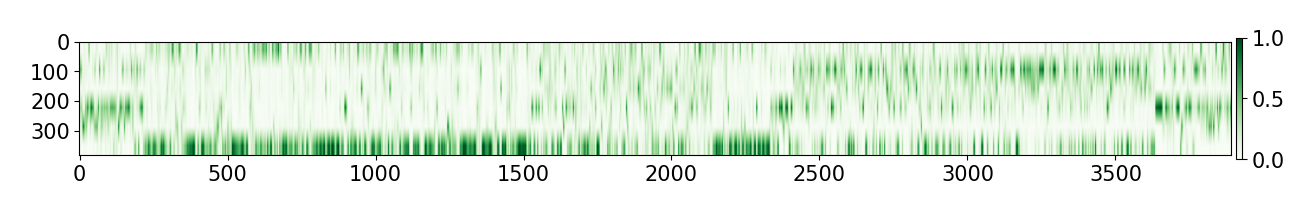}

   \includegraphics[width=\linewidth]{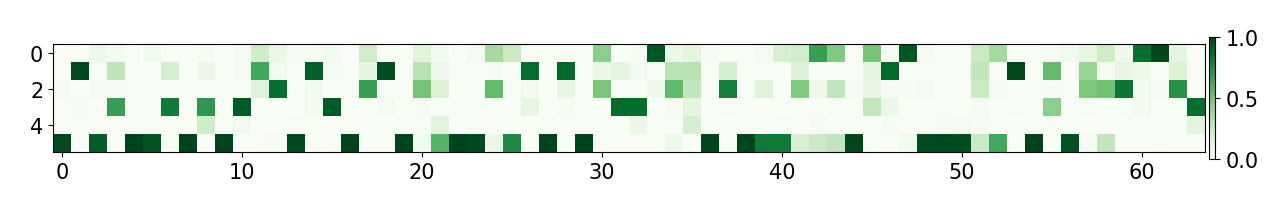}
   \caption{(top) Affinity between species classes (x-axis) from the S2 cells in Figure \ref{fig:s2_10k} and the average location embedding for each of the $6$ cells (y-axis). X-axis is scaled by $4$ pixels and y-axis by $32$ pixels for visibility. (bottom) Affinity between 64 expert centroids in the last transformer layer and S2 cell average location embedding.}
   \label{fig:s2_cell_class_affinity}
\end{figure}
\label{sec:geo_import}

\begin{figure}[ht]
  \centering
   \includegraphics[width=\linewidth]{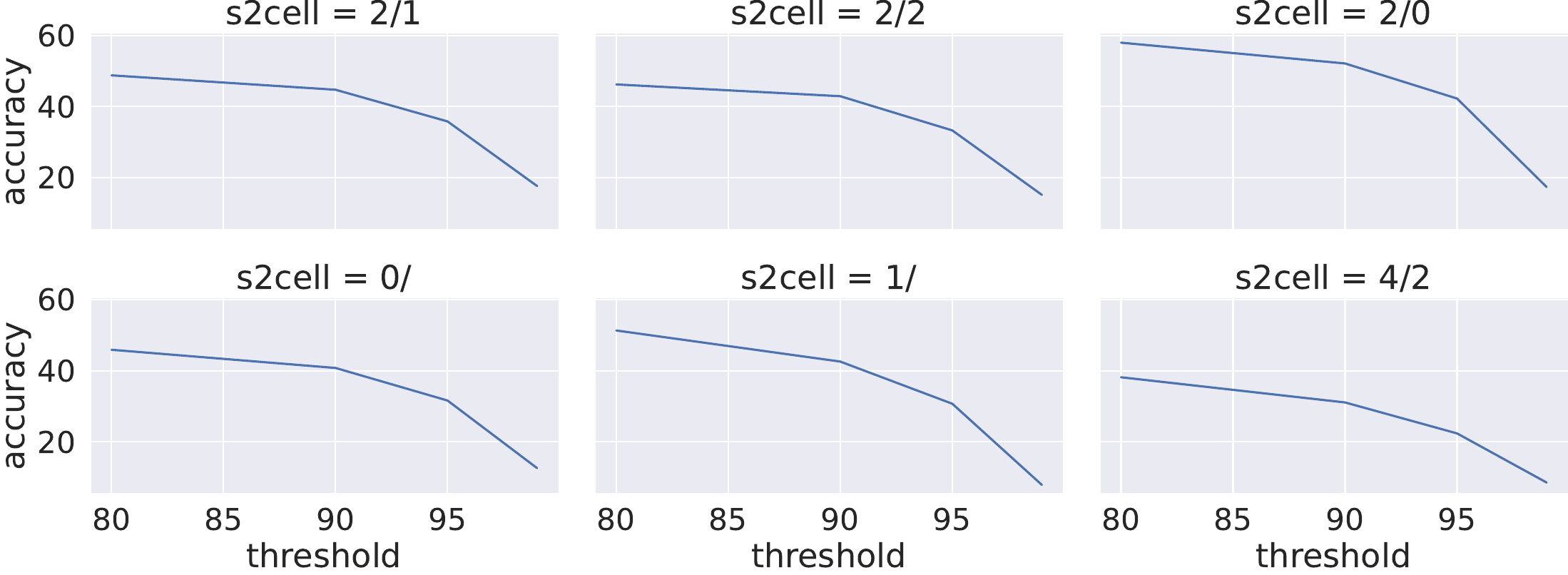}
   \\
   \vspace{0.3cm}
   \includegraphics[width=\linewidth]{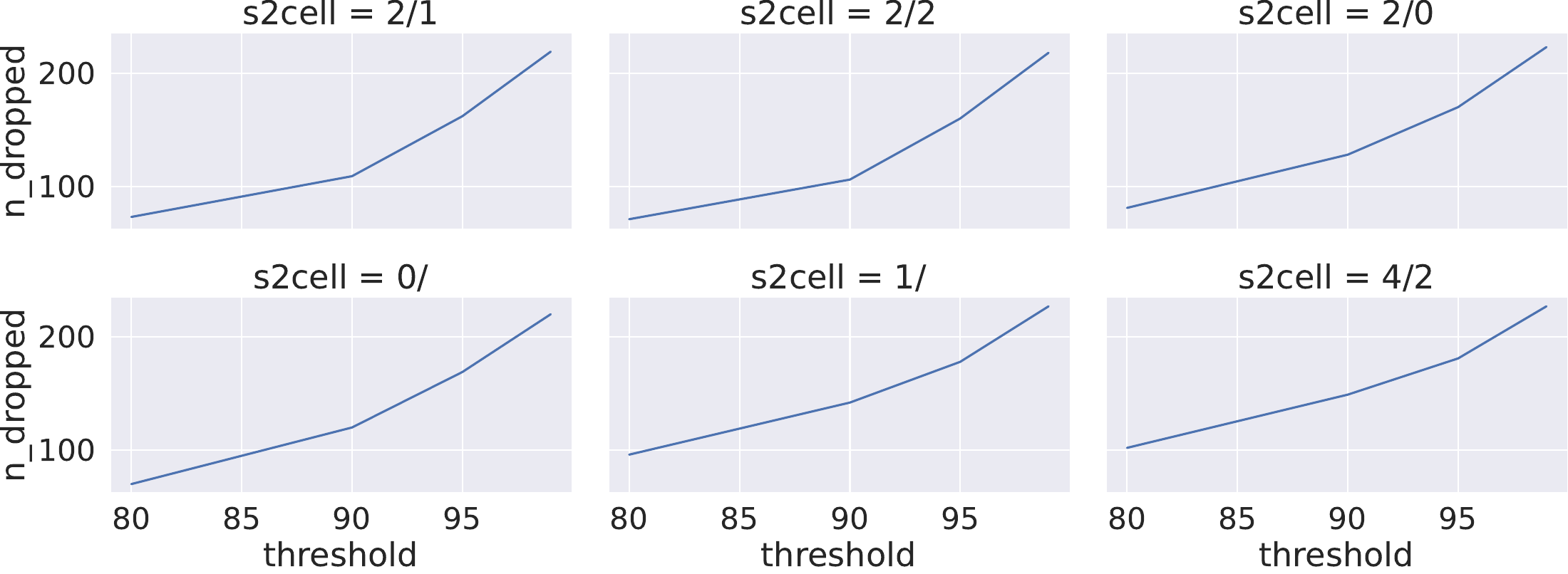}

   \caption{(top) Effect of dropping a set of experts on accuracy from a (non-finetuned) 64 expert model at layers $1,3,5,7$ and expert hidden MLP size of $2$ as selected by the percentile thresholding method described in Section \ref{sec:geo_import} on iNaturalist-Geo-10K S2 cells. (bottom) The effects of the same operations on the number of dropped experts out of $256$ experts ($64$ times $4$ layers). 
   }
   \label{fig:geo_exp_drop}
\end{figure}

\begin{figure}[!ht]
  \centering
   \includegraphics[width=0.5\linewidth]{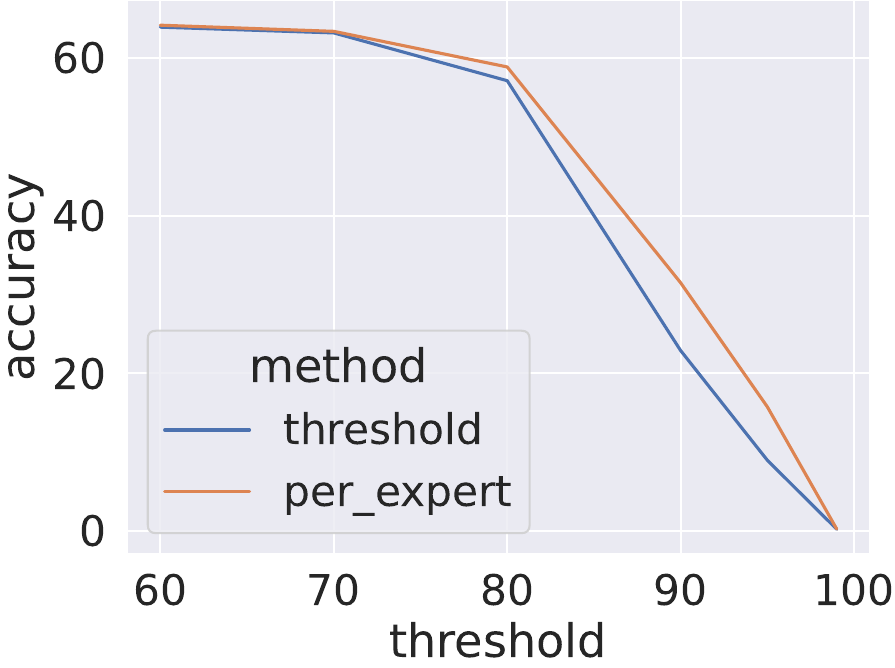}

   \caption{Comparing effects on accuracy of dropping experts by route popularity threshold or by per-expert importance for cell `2/1' on a location aware finetuned 16 expert model at layers $1,3,5,7$. 
   }
   \label{fig:exp_drop_method}
\end{figure}

\begin{table*}[!htbp]
    \centering
    \scriptsize
    \begin{tabular}{lrrrrrrrrrr} \\ 
    \hline
      Model & Finetune Data & Location & FLOPs (M) & Params (M) & L1 & L2 & L3 & L4 & L5 & L6  \\
     \hline
     MobileViTV2 & - & No & 468 & 2.94 & 67.10 & 66.91 & 64.63  & 64.50 & 61.02 & 70.54 \\
     & iNat-Geo-10K-6 & & & & \textbf{71.83} & \textit{69.12} & \textit{74.10} & \textit{67.06} &  \textbf{70.50} & \textbf{72.31} \\
     & iNat-Geo-10K-1 & & & & \textit{70.13} & \textbf{71.67} & \textbf{75.16} & \textbf{70.62} & 66.19 & \textit{70.90} \\
     MoE-MLP2 & - &  & \textbf{427} & \textbf{2.79} & 53.14 & 51.00 & 61.46 & 50.84 & 57.50 & 47.32 \\
     & iNat-Geo-10K-6 &   &  &  & 65.84 & 62.37 & 71.49 & 60.57 & 66.41 & 55.70 \\
     &  & Yes &  &  & 66.48 & 62.75 & 71.92 & 60.73 & \textit{66.97} & 55.70 \\
     & iNat-Geo-10K-1 & No &  &  & 62.49 & 59.23 & 67.01 & 56.59 & 60.58 & 49.76 \\
     \hline
     MobileViTV2 & - & No & 466 & 1.16 & 67.10 & 66.91 & 64.63  & 64.50 & 61.02 & 70.54 \\
      & iWildcam-6 &  & &  & \textbf{73.87} & \textbf{75.17} & \textbf{77.96} & \textit{81.54} & \textbf{93.85} & \textbf{75.20} \\
      & iWildcam-1 &  & &  & 69.18 & 66.36 & 63.61  & 38.83 & 69.75 & 64.54 \\
      MoE-MLP2 & - &  & \textbf{426} & \textbf{1.01} & 42.78 & 49.94 & 56.70  & 55.02 & 77.25 & 43.89 \\
       & iWildcam-6 &  & &  & 67.14 & 71.58 & 74.47  & 79.21 & 92.63 & 68.33 \\
        & & Yes & &  & \textit{70.99} & \textit{73.52} & \textit{77.43} & \textbf{81.90} & \textit{93.62} & \textit{72.53} \\
        & iWildcam-1 & No & &  & 62.37 & 47.65 & 46.51  & 46.53 & 42.87 & 46.94 \\
    
    \hline
    \end{tabular}
    \caption{Evaluation on location constrained data L1-L6 from iNaturalist cells and iWildcam clusters on the MoE models. All models are initialized with weights from MobileViTV2-0.5 pretrained on iNaturalist or WILDS-iWildcam. Finetuning data ending in `-6' filters training data to all 6 selected locations for the dataset. Data ending in `-1' finetunes the model only on data from the specific location used in testing the model. Empty data rows refer to no finetuning after pretrained weights are loaded (and experts model is initialized). MoE models have expert layers at $1,3,5,7$. MLP-$N$ models use $N$ hidden units in the expert MLP. `Location' indicates whether training is GPS aware.}
    \label{tab:geo_loc_models_ft}
\end{table*}

\subsection{Bias between Species and Geographical Location}
Figure \ref{fig:s2_cell_class_affinity} shows the affinity between classes and s2 cells by first computing the cosine distance between class patch embeddings and the layer projected geo encodings retrieved from TorchSpatial encoder ($F^{in}_{MLP}$ and $F_{loc}$ from Section \ref{sec:loc_enc}). We then use Softmax with temperature $0.001$ to generate an assignment probability distribution over cells for every patch and average over patches within the same class. As seen from Figure \ref{fig:s2_cell_class_affinity}, there appears to be biases between different species and S2 cells.  However, since we didn't utilize balanced sampling during training, this result does not account for the effect of class imbalance. Still, we see bias towards geographical regions by different experts, useful for structured pruning.

\subsection{Expert Dropping Behavior}
Accuracy generally decreases as more experts are dropped (shown in Figure \ref{fig:geo_exp_drop}) and is therefore a knob for applications that can trade off memory for accuracy. We compare expert dropping by computing threshold method from Section \ref{sec:threshold_dropping}, then after finding the number of experts to drop, we also drop the same number of experts with the lowest per-expert importance. As the threshold increases, the thresholding heuristic produces worse estimates as conveyed in Figure \ref{fig:exp_drop_method}. This is because, low popularity experts could still be useful in distinguishing foreground features as opposed to popular experts with background patch assigments. There is therefore a trade-off in compute time and accuracy between the two approaches. Computing per-expert importance scales as number of experts by time validation inference time while thresholding scales by number of threshold values by validation inference time. Since the per expert estimation is a one-time task, it might be tolerable if a model is not updated frequently.
 
\subsection{Geographical Expert Model Ablation}
From Table \ref{tab:geo_loc_models_ft}, we can see that location information provided performance improvement for the expert models for both datasets without any added parameters during inference. iWildcam benefited more from training with data from all clusters since it has less data and is more imbalanced. The location aware expert model reduced inference cost for iWildcam with minimal accuracy drop. However, on the larger iNaturalist-Geo-10K dataset, using the expert model with expert MLP hidden dimension of $2$ led to significant accuracy drops. It is therefore necessary to increase the MLP hidden dimension size at the expense of parameter count to recover lost accuracy as the dataset size increases. Finally, more iNaturalist cells benefited from finetuning with data from just the specific cell since it has more data compared to iWildcam.
\section{Conclusion}

In this work, we introduce the concept of conditional subnetworks for species recognition that are location aware. We then propose a training approach, a geographic subnetwork selection method and classification results on two species datasets. We demonstrate that even though species datasets tend to be imbalanced, we are still able to successfully incorporate different classes from different geographical regions in the subnetwork design. Some follow up directions of interest would include a) Exploring more fine grained geographical cell divisions to make stronger claims about geographical biases in experts. b) Utilizing concepts from the model pruning literature based on activations and weights (route popularity and a measure of expert MLP output activation magnitude) to close the gap between per expert and threshold expert dropping approach. c) Training on large scale camera trap datasets, e.g. LILA-BC camera trap API\footnote{\url{https://huggingface.co/datasets/society-ethics/lila_camera_traps}}, to make this set up of pre-trained geo subnetworks more accessible.






\bibliographystyle{named}

\end{document}